\documentclass[letterpaper, 10 pt, conference]{ieeeconf}  %

\IEEEoverridecommandlockouts                              %

\usepackage{colortbl}
\usepackage{graphicx}
\usepackage{hyperref}
\usepackage{cite}
\usepackage{amsmath}
\usepackage{amssymb}
\usepackage{bbold}
\usepackage{footnote}
\usepackage{xpatch}
\usepackage{float}
\usepackage{caption,subcaption}
\usepackage{flushend}
\usepackage[font=normalsize]{subcaption}
\usepackage[dvipsnames]{xcolor}
\definecolor{brightgreen}{rgb}{0.3, 0.8, 0.0}

\graphicspath{ {./images/} }
\newcommand{\bestresult}[1]{\boldmath{\textcolor{red}{#1}}}
\newcommand{\secondbest}[1]{\textcolor{blue}{\underline{#1}}}

\newcommand{\supdagger}{\textsuperscript{\textdagger}}
\newcommand{\etal}{\textit{et al.}}

\title{TransVisDrone: Spatio-Temporal \textbf{Trans}former for \textbf{Vis}ion-based \textbf{Drone}-to-Drone Detection in Aerial Videos}

\author{Tushar Sangam$^{1}$\supdagger, Ishan Rajendrakumar Dave$^{1}$, Waqas Sultani$^{2}$ and Mubarak Shah$^{1}$
\thanks{$^{1}$Center for Research in Computer Vision lab (CRCV), University of Central Florida, USA}
\thanks{$^{2}$Information Technology University of the Punjab, Lahore, Pakistan}
\thanks{\supdagger Corresponding Author: {\tt\small tusharsangam@knights.ucf.edu}}}

\date{August 2022}

\begin{document}
\pagestyle{plain} 

\maketitle
\thispagestyle{plain} 

\begin{abstract}

Drone-to-drone detection using visual feed has crucial applications, such as detecting drone collisions, detecting drone attacks, or coordinating flight with other drones. However, existing methods are computationally costly, follow non-end-to-end optimization, and have complex multi-stage pipelines, making them less suitable for real-time deployment on edge devices. In this work, we propose a simple yet effective framework, \textit{TransVisDrone}, that provides an end-to-end solution with higher computational efficiency. We utilize CSPDarkNet-53 network to learn object-related spatial features and VideoSwin model to improve drone detection in challenging scenarios by learning spatio-temporal dependencies of drone motion. Our method achieves state-of-the-art performance on three challenging real-world datasets (Average Precision@0.5IOU): NPS 0.95, FLDrones 0.75, and AOT 0.80, and a higher throughput than previous methods. We also demonstrate its deployment capability on edge devices and its usefulness in detecting drone-collision (encounter). Project: \url{https://tusharsangam.github.io/TransVisDrone-project-page/}

\end{abstract}

\section{Introduction}

Drones have seen great popularity in various real-world applications such as surveillance, package delivery, military applications, agricultural robotics~\cite{king2017technology}, etc. From the perception (computer vision) point of view, drone visual feed can be used to address various problems such as human action recognition~\cite{sultani2021human, timebalance, gabv1, gabv2, spact}, behavioral understanding~\cite{li2021uav}, ground object detection and tracking~\cite{aakash_iros22, cao2021visdrone}, etc. Apart from  %
the above ground object/actors detection, it is also crucial to detect other airborne objects like the other drones or birds to prevent collisions during flight.~\cite{rozantsev2016detecting}, tackle a drone attack~\cite{jacobsen2021security}, or coordinate flights with other drones.\cite{hassija2021fast}. Although drone detection from aerial videos has crucial prospects, it is an under-explored research problem. 

Drone-to-Drone detection has a more challenging nature compared to standard object detection problems. The major challenges are:  First, \textit{small-sized  object (drone)}: in the captured aerial videos, typical target drones are  only 0.07\% of the frame-size~\cite{li2016multi:conf_typical}, whereas in standard detection problems, the object size is  about 20\%~\cite{deng2009imagenet, coco}. Second,  \textit{Movement of target drones} can be fast and erratic which often blurs video frames and make it  difficult to detect objects. Third,  \textit{Egomotion} or the source drone movement also makes the detection and tracking of the target drone difficult.

\begin{figure}[H]

    \begin{subfigure}{0.5\textwidth}
        \vspace{2mm}
        \centering
        \includegraphics[width=0.8\textwidth]{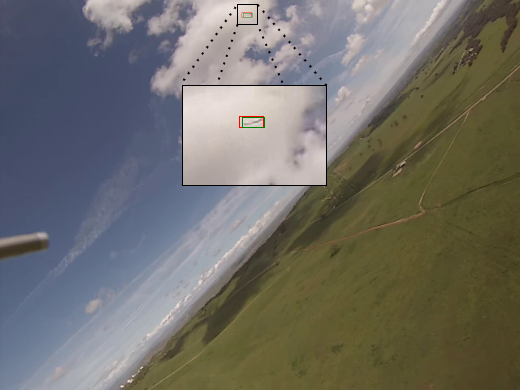}
        \vspace{-2mm}
        \caption{\footnotesize Target-drone is above-the-horizon with the clouds in its background. \label{fig:nps_1}}

    \end{subfigure}
    
    \begin{subfigure}{0.5\textwidth}
        \vspace{1mm}
        \centering
        \includegraphics[width=0.8\textwidth]{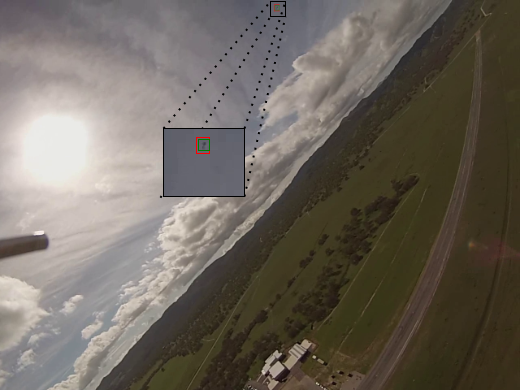}
        \vspace{-2mm}
        \caption{\footnotesize Abrupt illumination change due to egomotion and facing the sun. \label{fig:nps_2}}

    \end{subfigure}

    \hfill
    \begin{subfigure}{0.5\textwidth}
    \vspace{2mm}
    \centering
        \includegraphics[width=0.8\textwidth, height=2.2in]{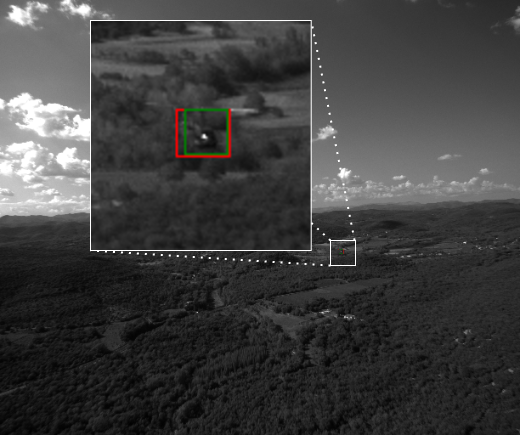}
        \vspace{-2mm}
        \caption{\footnotesize Target-drone is below-the-horizon which results in a cluttered\\
        background with movements in vegetation or other ground vehicles.\label{fig:aot_1}}

    \end{subfigure}
    \hfill
    \vspace{-2mm}
    \caption{\small{\textbf{Qualitative Visualization}. Our TransVisDrone method is successfully able to detect the drone even in various challenging scenarios. \textcolor{brightgreen}{Green Box} indicates ground-truth, \textcolor{red}{Red Box} indicates output prediction box.} (Best view with 300\% zoom-in)}
    \label{fig:visualize}
\end{figure}
Fourth, \textit{Uncontrolled Surroundings}: aerial videos are often captured in an outdoor setting, which brings many challenges like variable lighting, cluttered background, occlusion, etc. Apart from these challenges, computational efficiency (i.e. throughput) is highly desirable since most drones need to process the data on edge-computing resources.

There are a few standard object detection methods such as  Mask-RCNN~\cite{maskrcnn}, YOLO~\cite{yolo}, and De-DETR~\cite{dedetr} which can be employed to detect drones in a drone video.
However, since they do not exploit the temporal dynamics of the video, they perform at a sub-optimal level due to a lack of  temporal coherence in the predictions and can miss the target drone when there is motion due to source or target drone or both. 

A recent method \textit{DogFight}~\cite{Ashraf_2021_CVPR:conf_typical} utilizes a two-stage segmentation approach. As a part of preprocessing, a frame is first divided into overlapping patches. In the first stage, contextual information  in each frame is learned through a 2-D convolution network and channel-pixel-wise attention. The frame-wise detections obtained through the first stage are connected through a connected component analysis and an off-the-shelf tracker. These obtained tracks are processed through 3D-Convolution networks and channel-pixel-wise attention. Although~\cite{Ashraf_2021_CVPR:conf_typical} achieves the state of art  results, %
it has several downsides: First, \textit{High computational requirement}: since their framework utilizes 2D and 3D convolution networks along with channel-pixel attention modules in both stages, it requires large computational resources. Second,
\textit{Low throughput/ FPS}: %
the framework takes multiple overlapping crops from a single frame and processes them individually. It also utilizes connected component analysis and an off-the-shelf tracker which are implemented on CPU and do not get benefit from the parallel operations of GPU. Third, \textit{Non-Differential Components}: It is not an end-end approach; i.e.  between stage-1 and stage-2, the method requires non-differentiable components like connected component analysis and off-the-shelf tracker, which makes it complicated to train, since  first, we need to store tracks from stage-1 for all videos and then start training stage-2. The non-differentiable components also require hand-crafted filters which introduce dataset-specific inductive biases. Rozantsev~\etal~\cite{rozantsev2016detecting} also face similar issues to ~\cite{Ashraf_2021_CVPR:conf_typical}.

In the spirit of tackling the challenges of~\cite{Ashraf_2021_CVPR:conf_typical,rozantsev2016detecting} and other prior work for real-time applications, we propose a  new framework: spatio-temporal TRANSformer for VISion-based DRONE-to-Drone Detection (\textit{TransVisDrone}). We propose a simple end-to-end framework based on CSP-DarkNet53~\cite{wang2020cspnet} and Video-Swin transformer~\cite{videoswin}. Our overall framework  is shown in Fig.~\ref{fig:framework}. First, a clip is sampled from the flight video within a temporal window and processed through stochastic temporally-consistent transformations. The transformed clip is flattened across time dimension and fed to CSP-DarkNet53 to obtain spatial features in each frame. Finally, to exploit video temporal information, 
drone features of a short video clip are passed through Video-Swin model to learn the spatio-temporal dependencies among them. The final output is obtained through standard detection head and non-maximum suppression operations. 

The major contributions of our work can be summarized as follows:
\begin{itemize}
    \item We propose a simple, efficient, and end-to-end trainable framework for drone detection in the videos captured from a drone. Our approach detects drones using only video feed  without relying on expensive payloads, Lidars, etc, ~\cite{8794243,9691910}. Our method is fully differentiable and does not require any handcrafted algorithms as used in the prior work \cite{Ashraf_2021_CVPR:conf_typical,rozantsev2016detecting:article_typical}.
    
    \item Our method establishes new state-of-the-art on three publicly available drone detection datasets: NPS~\cite{li2016multi:conf_typical}, FL-Drone~\cite{rozantsev2016detecting:article_typical}, and  AOT dataset~\cite{AOT:IEEEwebsite}. These datasets differ significantly in complexity, drone sizes, and a number of videos. 
    
    \item We perform detailed ablation studies of various design choices of our framework and conduct experiments on edge-device to demonstrate the usefulness of our approach for real-world applications like drone encounter detection. 
\end{itemize}

\section {Related Works}

\subsection{Drone detection}
Drone detection has been studied by various works in the aerial robotics community, however, they mainly involve non-visual sensor data or static cameras from the ground. Dressel and Kochenderfer~\cite{8794243} try to detect other intruder drones for security purposes using RF sensors of the target drones. Since this approach is constrained to the attached sensors of the target drones, it is not applicable to airborne objects that do not bear the RF sensors such as birds and balloons. 
~\cite{nava2022learning} proposes a self-supervised learning approach for visual localization of a quadrotor using its own noise as a source of supervision. ~\cite{li2022self} presents an efficient self-supervised deep neural network approach for monocular multi-robot relative localization. Yang and Quan~\cite{inproceedings} use visual feed and computer vision models to guide the servo and intercept intruder drones. However, their data is collected from a static camera on the ground in a controlled environment and it only captures low-altitude samples. Since their image acquisition setup is highly controlled, it is not suitable for the drone camera and detecting challenging surroundings. 
Chen~\etal~\cite{9561365} uses point cloud data to segment the voxels and avoid obstacles. However, obtaining the point cloud data is not inexpensive and requires LIDAR sensors.
Dogru and Marques~\cite{9691910} also use LIDAR sensors placed on the ground to detect UAVs in the air. 
Cao~\etal~\cite{unknown} use the Siamese network to perform visual tracking of the objects in the UAV camera feed. However, they focus on the objects on the ground instead of other airborne objects. Also, they only process the spatial information and do not take the temporal context into consideration.
Yu~\etal~\cite{9341432} uses visual sensors along with radars to avoid obstacles. However, their visual computing is based on optical flow which is computationally slow for a real-time system. Also, their work is not demonstrated on real-world datasets.
Rozantsev~\etal~\cite{rozantsev2016detecting:article_typical} propose a sptio-temporal(ST) cube that can combine the spatial \& motion features. To achieve that they employ two CNNs in a sliding window fashion \& followed by a third CNN to detect UAVs in each ST cube. This approach suffers from the similar problems of being multi-stage, complex post-processing, and computationally expensive to deploy.

\begin{figure*}
\begin{center}
  \includegraphics[width=0.83\textwidth]{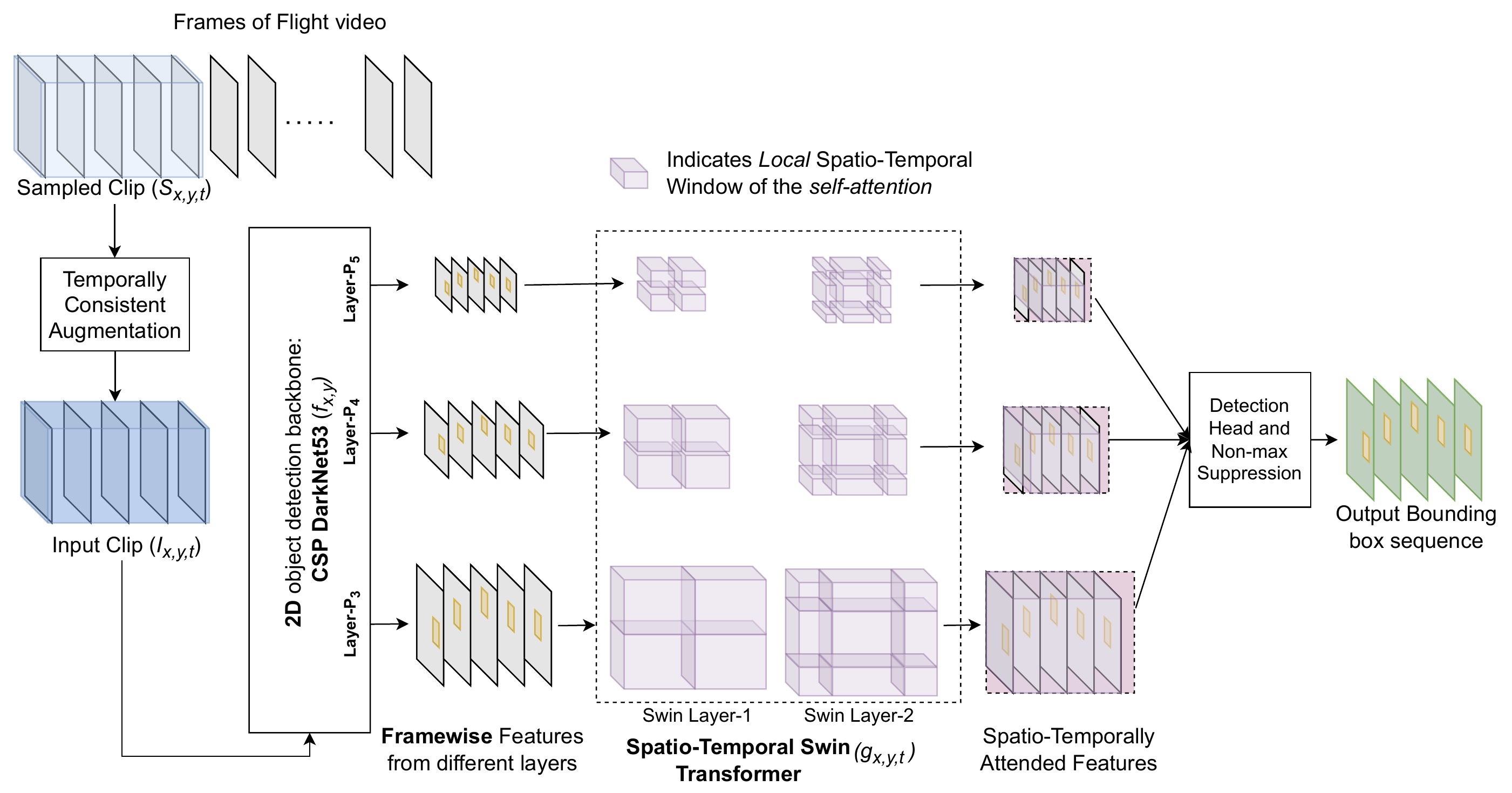}
\end{center}
\vspace{-2mm}
\caption{\textbf{TransVisDrone Framework}: First, a set of frames is sampled within a temporal window (shown in blue color) from the captured flight video into a clip. This sampled clip is transformed through temporal-consistent augmentations to obtain the input clip for our detection model  $I_{x,y,t}$, here $x,y$ indicates pixel coordinates, and $t$ indicates the timestamp of the frame. Each frame of input clip is passed through CSP-DarkNet53 ($f_{x,y}$)~\cite{wang2020cspnet} to obtain object-related \textit{spatial} features from different layers ($P_{3}, P_{4}$, and $P_{5}$). To learn the \textit{spatio-temporal dependencies} from these frame-wise spatial features, we use \textit{Spatio-Temporal Swin Transformer}($g_{x,y,t}$)~\cite{videoswin}. Finally, output tokens of the transformer are sent to the detection head and non-maximum suppression module to obtain the final detection output object bounding box sequence.}
\label{fig:framework}
\vspace{-3mm}

\end{figure*}

\subsection{Transformers for temporal context}
We want to improve drone detection in videos by learning drone motion from the temporal context. Recently, transformer-based self-attention is gaining popularity in the computer vision community because of its ability to focus on a non-local affinity of the data, which was previously a major limitation of convolution neural networks. A self-attention can successfully learn global temporal context from the sequence of frames and can encode the appearance and motion information of the moving object~\cite{vtn}. However, learning the global temporal context via transformer is computationally expensive, hence, Liu~\etal~\cite{swin, videoswin} develop a method to cut the cost of global self-attention by introducing a shifted window-based attention mechanism. Considering the ability to deploy on edge-computing devices, we choose VideoSwin instead of a full self-attention-based transformer.

\section{Method}\label{method}
The proposed method is based on the following three insights: First, to tackle the large variation of drones and background scenes, data augmentation which maintains the video information in a short clip creates challenging scenarios; Second,  due to real time applications and tiny object sizes, a fast and multi-scale, multi-level features extractor should be used for accurate detections; Third, temporal (video) information should be exploited while attending the important regions in the videos. To accomplish these goals, our framework consists of three components: (1) Temporally consistent preprocessing, (2) Spatial feature extractor module
(3) Spatio-temporal SwinTransformer~\cite{videoswin} module. A schematic diagram of our framework is shown in Fig.~\ref{fig:framework}.%

\subsection{Preprocessing}\label{TCA}
From any $i^{th}$ flight video $V^{i}$, frames within a temporal window, $\tau$ are selected, which we call sampled clip $S_{x,y,t}$. In the training mode, we select the sampled clip from a random temporal location, whereas, during the inference mode, we use the sliding window method to cover the whole video. The sampled clip is then transformed through Temporally-Consistent Augmentation (TCA) module. The main goal of this module is to first select a random subset of augmentations (details in Sec.~\ref{sec:implementation})
and apply the same augmentations on every frame of $S_{x,y,t}$. Note that, if each frame of the clip is transformed differently, it destroys the temporal dynamics of video~\cite{tclr, svt}. %

\subsection{Spatial feature module}\label{backbone}

In CNN, feature map resolution is decreased in subsequent layers using max-pooling operation which retains only the maximum values in local regions, resulting in the loss of fine geometrical details required for small object detection.  
Thus we use CSPDarknet53~\cite{yolov4} which doesn't downscale features by the common max-pooling operation \& produces multi-scale features using \textbf{s}patial \textbf{p}yramid \textbf{p}ooling block SPP~\cite{spp}.
Each frame sampled at the end of the TCA~\ref{TCA} module is then passed through the CSPDarknet53~\cite{yolov4} backbone and we obtain multi-scale ($P_{3}, P_{4}, P_{5}$) spatial features for each frame in the clip samples.

\subsection{Spatio-Temporal Transformer}\label{method:swin}
Spatial features obtained in Sec.~\ref{backbone} lack temporal video information and often miss drones due to their abrupt shape change and motion blur.
Therefore, these multi-scale spatial features are %
fed to corresponding VideoSwin~\cite{videoswin} branches to learn spatio-temporal dependencies. Figure~\ref{fig:framework} shows three different VideoSwin~\cite{videoswin} branches being applied at three different feature scales.
The standard \textbf{M}ulti \textbf{H}ead \textbf{S}elf \textbf{A}ttention MSA~\cite{NIPS2017_3f5ee243} layer works by dividing the input into fixed-sized small patches \& then learning their relations through self-attention.   
SwinTransformer~\cite{videoswin} blocks work more efficiently than the standard MSA head\cite{NIPS2017_3f5ee243} by dividing the data into bigger patch sizes.
In our case, we set the default patch size of $8\times8 \times \tau$. If the resolution of the spatial feature is $H \times W$, then the input feature size of the VideoSwin branch will be $H\times W \times \tau$. Thus after patching the input spatial feature with $\frac{H}{8} \times \frac{W}{8}$ patches and each patch size becomes $8 \times 8 \times \tau$. It then applies the MSA~\cite{NIPS2017_3f5ee243} inside each 3D patch, by further dividing this patch into smaller $M \times M \times \tau$  patches. This is the first layer of the SwinTransformer~\cite{videoswin} block referred to as \textit{3DW-MSA} (3D Window-MSA) layer. For the next layer, windows are shifted by $(\frac{8}{2} ,\frac{8}{2}, 0)$ and MSA~\cite{NIPS2017_3f5ee243} is applied again. This layer is referred to as \textit{3DSW-MSA} (3D Shifted Window-MSA). Since there is a spatial overlap between layer 1 \& layer 2, it can capture the spatio-temporal global cues over the large receptive field effectively. Since there is a relatively small translation motion in the consecutive frames, attending over the local area instead of the entire spatial map is cost-effective \& performant. Details of the design choice of the attention window size are analyzed in Sec~\ref{sec:videoswin}.

\subsection{Loss functions}
To optimize our framework, we utilize standard loss functions introduced by YOLO~\cite{yolo}, which are: (1) objectness loss (2) classification loss, and (3) localization loss. The feature map $g(f(I_{x,y,t}))$ is considered as a $S\times S$ grid, and in each cell, $B$ bounding boxes are predicted and prediction losses are applied. As shown in the following equation, the objectness loss is calculated based on the condition if the object is present or not in the cell.
\begin{multline}
\small
\mathcal{L}_{objectness} = \sum_{k = 0}^{S^2}
    \sum_{j = 0}^{B}
        {\mathbb{1}}_{kj}^{\text{obj}}
        \left(
            C_k - \hat{C}_k
        \right)^2 \\
        + \lambda_\textrm{noobj}
\sum_{k = 0}^{S^2}
    \sum_{j = 0}^{B}
    ({1-\mathbb{1}}_{kj}^{\text{obj}})
        \left(
            C_k - \hat{C}_k
        \right)^2,
\label{eq:objness}
\end{multline}
\normalsize
where, ${\mathbb{1}}_{kj}^{\text{obj}}$ is indicator binary function which takes value of 1 if $j^{th}$ bounding box in cell k contains the object. $\lambda_\textrm{noobj}$ is a hyperparameter which is set to 5 as per~\cite{yolo}. As shown in the Equation 2, a class-specific loss is computed using square of the error between predicted conditional class probability $\hat{p}_k(c)$ and ground-truth $p_k(c)$ for cell $k$. 
\begin{equation}
\mathcal{L}_{classification} = \sum_{k = 0}^{S^2}\mathbb{1}_{k}^{\text{obj}}\sum_{c \in \textrm{classes}}\left(p_k(c) - \hat{p}_k(c)\right)^2,
\end{equation}
where, $\mathbb{1}_{k}^{\text{obj}}$ is indicator binary function which takes value of 1 if an object is present in the $k^{th}$ cell. The third loss is computed from the discrepancy between the predicted bounding box and the ground-truth as shown in the equation below.
\vspace{-1mm}
\small
\begin{multline}
    \mathcal{L}_{localization} = \sum_{k = 0}^{S^2}
        \sum_{j = 0}^{B}
         {\mathbb{1}}_{kj}^{\text{obj}}
                \left[
                \left(
                    x_k - \hat{x}_k
                \right)^2 +
                \left(
                    y_k - \hat{y}_k
                \right)^2
                \right]
    \\
    + \lambda_\textbf{coord} 
    \sum_{k = 0}^{S^2}
        \sum_{j = 0}^{B}
            {\mathbb{1}}_{kj}^{\text{obj}}
             \left[
            \left(
                \sqrt{w_k} - \sqrt{\hat{w}_k}
            \right)^2 +
            \left(
                \sqrt{h_k} - \sqrt{\hat{h}_k}
            \right)^2\right],
\end{multline}
\normalsize
where, ($\hat{x}_{k}$, $\hat{y}_{k}$) and  (${x}_{k}$, ${y}_{k}$) are predicted top-left corner coordinates of predicted and groundtruth bounding box, respectively. Whereas, ($\hat{w}_{k}$, $\hat{w}_{k}$) and  (${h}_{k}$, ${h}_{k}$) are predicted width and height of predicted and groundtruth bounding box, respectively. $\lambda_\textbf{coord}$ is a hyperparameter set to 5 as per~\cite{yolo}.
\section{Experiments}

\subsection{Implementation Details}
\label{sec:implementation}

Since target drones are not always present in every frame of flight videos,  drone datasets~\cite{li2016multi:conf_typical, rozantsev2016detecting:article_typical, AOT:IEEEwebsite} do not provide any annotations for such empty frames. Therefore, for training, we use only frames that provide the annotations, and for evaluation, we test all frames with a  skip rate of 4 following the protocols of prior work~\cite{Ashraf_2021_CVPR:conf_typical}. 
In Temporal Consistent Augmentation module, we use standard augmentations like perspective transforms, cutout, and color jittering. In order to reduce the extensive hyperparameter search for the augmentations, we follow the augmentation strengths of Zhu~\etal~\cite{Zhu_2021_ICCV} which deals with small ground-object detection. After augmentation module, we resize the frame to $1920\times1280$. We use Adam \cite{kingma2014adam:article_typical} optimizer with a momentum of 0.843 \& learning rate of 3e-5. We apply cosine-decay learning rate scheduler~\cite{https://doi.org/10.48550/arxiv.1812.01187} during training. Following the prior work~\cite{Ashraf_2021_CVPR:conf_typical}, we start the training from publicly available pretrained model weights of yolov5l~\cite{glennjocher20204154370} on MS-COCO~\cite{lin2014microsoft:conf_typical}. For the non-maximum suppression module, we set IoU threshold at 0.6 and the confidence threshold at 0.001. Our codebase can be found on GitHub \footnote{\href{https://github.com/tusharsangam/TransVisDrone}{https://github.com/tusharsangam/TransVisDrone}}.

\subsection{Evaluation Protocol}\label{evalautionsection}
For evaluation, we set the IoU-threshold between predictions and ground truth to 0.5. Therefore detections matching with ground-truth with IoU$\geq$0.5 are counted as the true positives. Following the prior work, we report average precision (AP) and a Precision-Recall pair corresponding to the best F1-Score. Since AP is averaged over uniformly spaced 11 operating points of the precision-recall curve, it is more reliable compared to the best precision-recall pair.  Following prior work~\cite{Ashraf_2021_CVPR:conf_typical}, we evaluate on every $4^{th}$ frame.

\subsection{Datasets}
    We  use three challenging drone datasets.
    Following Dogfight \cite{Ashraf_2021_CVPR:conf_typical}, we use NPS \cite{li2016multi:conf_typical} and Fl-drones \cite{rozantsev2016detecting:article_typical}. In addition,  we also provide evaluations on the newly released Airborne Object Tracking Dataset (AOT)~\cite{AOT:IEEEwebsite}%
    ~\cite{ICCVWorkshop:IEEEwebsite}.

    \noindent\textbf{NPS-Drones dataset \cite{li2016multi:conf_typical}}
This dataset has 50 videos of custom delta wing air frames with a total number of frames adding up to 70250. Videos are captured from GoPro 3 camera with an HD resolution of $1920\times1280$ or $1280\times760$.
The resolution is sufficient for capturing objects at a far distance range. Objects in this dataset are mainly small drones as shown in Fig.~\ref{fig:nps_1} \& Fig.~\ref{fig:nps_2}. 
    Object size typically ranges from $10\times8$ to $65\times21$. We use the clean version annotations released by Ashraf~\etal~\cite{Ashraf_2021_CVPR:conf_typical}. 
    Following the train/val/test split of~\cite{Ashraf_2021_CVPR:conf_typical}, we use video-id \#01-\#36 for training, videos \#37-\#40 for validation and \#41-\#50 videos for testing split. 
    \newline
    \textbf{FL-Drones dataset \cite{rozantsev2016detecting:article_typical}} This small-scale dataset is proposed by EPFL with 14 videos which totals up to 38,948 frames. It is captured from a camera mounted on the flying drones with a mix of indoor and outdoor scenes.
    Frames are at the resolution of $640\times480$ or $752\times480$ in grayscale. 
    Additionally drone object sizes vary from $9\times9$ to $259\times197$.
    Following Ashraf~\etal~\cite{Ashraf_2021_CVPR:conf_typical} we use half of the frames of each video for training and the other half for testing. We use the cleaned version of annotations released by Ashraf~\etal~\cite{Ashraf_2021_CVPR:conf_typical}.

    \noindent \textbf{Airborne Object Tracking AOT dataset\cite{AOT:IEEEwebsite}}
    This was released for the ICCV 2021 workshop challenge \cite{ICCVWorkshop:IEEEwebsite} hosted by \textit{Amazon Prime Air}. This dataset is collected from a high-resolution camera mounted onboard aerial vehicles. It contains up to 5.9M+ frames, collected at the resolution of $2448\times2048$ in grayscale. The planned encounters in the flight sequences have trajectories that are intended to generate a broad range of distances, closing speeds, and angles of approach. Airborne objects include helicopters, airplanes, drones, and other unplanned airborne objects such as birds, flocks, and balloons. The objects have labels, but information about their distance is not provided. Refer to Fig.~\ref{fig:aot_1} for a sample frame from this dataset. The original challenge test set is sequestered and the evaluation server is no longer active. Since there is no test set released, we use part\#1 of the dataset where 516 videos are for training \& 171 videos for testing, while the remaining 300 videos are for validation. 

\subsection{Comparison with Prior works}    
\begin{figure}
\vspace{2mm}
\begin{center}
  \includegraphics[trim={0.5cm 0cm 1.5cm 1.3cm},clip, width=0.70\columnwidth]{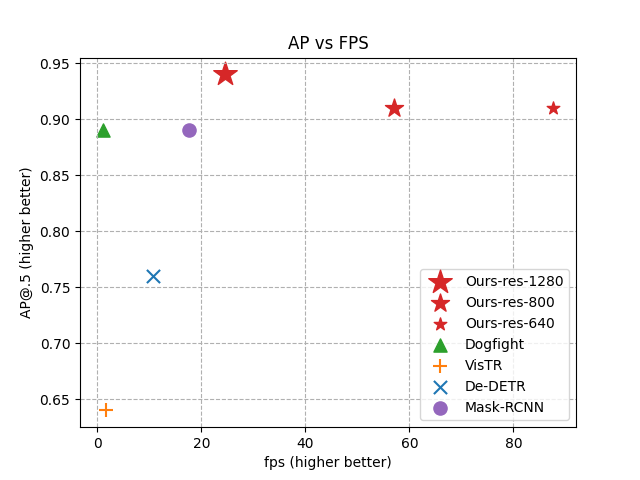}
\end{center}
\vspace{-4mm}
\caption{\textbf{Tradeoff for performance AP@0.5IoU vs Throughput (FPS)}: Our method TransVisDrone consistently outperforms prior works under different input resolutions. Other methods are evaluated on 1280 resolution. No off-the-shelf optimizations.}
\label{fig:APvfps}
\vspace{-5mm}

\end{figure}

We compare our method with several recent state-of-art methods on NPS and FLDrones datasets as shown in Table~\ref{tab:NPS_SOTA_table}. Performance wise our TransVisDrone outperforms the prior method by \textbf{3\%} and \textbf{1\%} absolute on \textbf{AP} metric on \textbf{NPS}~\cite{li2016multi:conf_typical} and \textbf{FLDrone datasets}~\cite{rozantsev2016detecting:article_typical}, respectively. As shown in Fig.~\ref{fig:APvfps} and Table~\ref{tab:NPS_SOTA_table}, with the comparable performance, our method is significantly better than prior methods in terms of throughput which is measured in frames-per-second. Experiments performed on NVIDIA RTX A6000 48G GPU.

\begin{figure}[h]
 \vspace{-1mm}
\begin{center}
  \includegraphics[width=0.50\columnwidth]{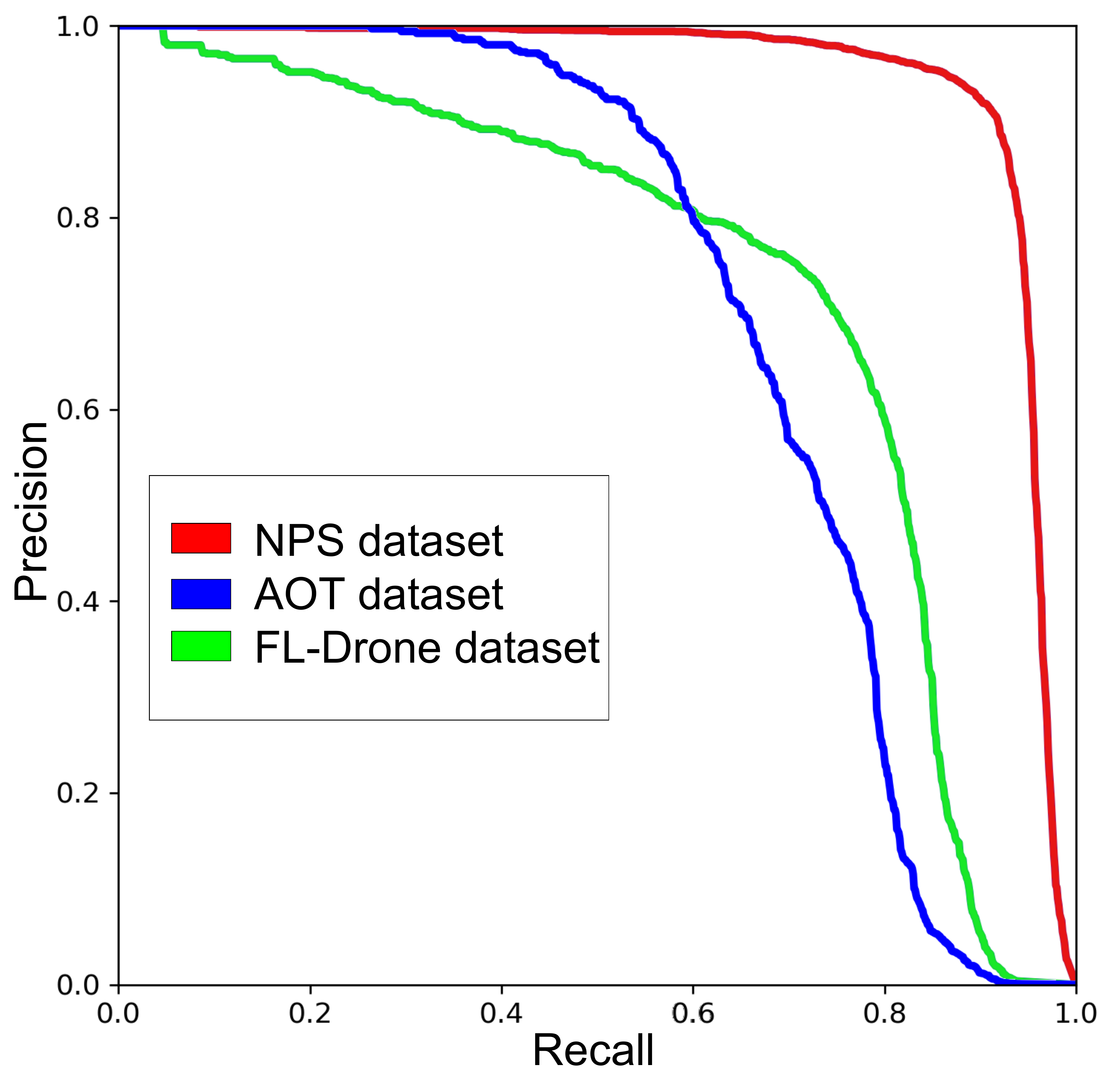}
\end{center}
\vspace{-4mm}
\caption{\textbf{Precision vs Recall Curve} of our method on there benchmark datasets.}
\label{fig:roc}
\vspace{-3mm}
\end{figure}

We also propose evaluation on AOT dataset~\cite{AOT:IEEEwebsite}, where we show results of our method and train a new model of Dogfight~\cite{Ashraf_2021_CVPR:conf_typical} using their official code repository. As shown in Table~\ref{tab:aot}, we outperform the prior method by \textbf{6\%} absolute on AOT dataset~\cite{AOT:IEEEwebsite}. Various operating points of our method can be seen at the precision-recall curve in Fig.~\ref{fig:roc}. 
\begin{table}[h]
\centering
\begin{tabular}{lccc}
\hline

\hline

\hline\\[-3mm]
\textbf{Methods} & \textbf{AP-NPS~\cite{li2016multi:conf_typical}}    &\textbf{AP-FLDrones~\cite{rozantsev2016detecting:article_typical}}  & \textbf{FPS}\\ 
\hline

SCRDet-H~\cite{scrdet}         & 0.65                 & 0.52   &      \\
SCRDet-R~\cite{scrdet}         & 0.61                 & 0.52   &      \\
FCOS~\cite{fcos}             & 0.83                 & 0.62   &     \\
Mask-RCNN~\cite{maskrcnn}        & 0.89                 & 0.68   &   17.55   \\
MEGA~\cite{mega}             & 0.83                 & 0.65   &      \\
SLSA~\cite{slsa}             & 0.46                 & 0.61   &      \\
De-DETR~\cite{dedetr}          & 0.76                 & \_     &  10.69 \\
VisTR~\cite{vistr}            & 0.66                 &  \_    & 1.6   \\
yolov5-tph~\cite{Zhu_2021_ICCV}       & \secondbest{0.92}    &  0.69 & 25    \\
Dogfight~\cite{Ashraf_2021_CVPR:conf_typical}         & 0.89              & \secondbest{0.72}  & 1.0\\
\textbf{Ours}    & \bestresult{0.95}   & \bestresult{0.75}   & \textbf{24.6}\\
\hline

\hline

\hline\\[-3mm]
\end{tabular}
\caption{\label{tab:NPS_SOTA_table}
\textbf{Drone-to-drone detection results on NPS \cite{li2016multi:conf_typical} and FL-drones \cite{rozantsev2016detecting:article_typical} datasets}. The best method is shown in \bestresult{Red} and the second best method is shown in \secondbest{Blue}.} 
\vspace{-3mm}
\end{table}

\begin{table}[h]
\vspace{-1mm}
\centering
\begin{tabular}{lccc} 
\hline

\hline

\hline\\[-3mm]
\textbf{Methods} & \textbf{AP}  & \textbf{Precision} & \textbf{Recall}  \\ 
\hline

Dogfight~\cite{Ashraf_2021_CVPR:conf_typical}         & {0.74} & {0.82} & {0.65} \\
\textbf{Ours}    & {\textbf{0.80}} & \textbf{0.82} & \textbf{0.72 }\\
\hline

\hline

\hline\\[-3mm]
\end{tabular}
\caption{
\textbf{Drone-to-drone detection results on AOT~\cite{AOT:IEEEwebsite}.}}
\label{tab:aot}
\vspace{-5mm}

\end{table}

\vspace{-3mm}
\subsection{Ablations}
In this section, we study various building blocks and design choices of our framework and try to report their performance and throughput trade-off. 
\subsubsection{Size of Temporal Window} In order to understand the impact of temporal context in our framework, we carry out experiments with three different sizes of the temporal window ($\tau$): \{1,3,5\}. Experiments are performed on NPS and AOT datasets with 1280 resolution. We do not perform $\tau=5$ on AOT due to its immense training time. As shown in Table~\ref{table:num-of-frames}, increasing the temporal window size from 1 to 3, gets noticeable performance improvement, whereas, further increasing the size to 5  slightly improves the precision and recall. This implies that short-term motion ($\tau=3$) is helpful in is very helpful in learning the spatio-temporal features of drone motion which results in improving the detection performance over single-frame detection, but the improvement diminishes with longer temporal context. %
In our final setting, we use $\tau=5$ due to its better performance on precision-recall.

\begin{table}[h]
\vspace{2mm}
\centering
\begin{tabular}{lccc} 
\hline

\hline

\hline\\[-3mm]
\textbf{Our model on NPS \cite{li2016multi:conf_typical}} & \textbf{AP}  & \textbf{Precision} & \textbf{Recall}  \\ 
\hline
$\tau = 1$         & 0.93          & 0.88               & 0.89             \\
$\tau = 3$          & {0.95}             & {0.91} & {0.90}         \\
$\tau = 5$          & {\textbf{0.95}}          & {\textbf{0.92}} & {\textbf{0.91}}         \\

\hline
\textbf{Our model on AOT \cite{AOT:IEEEwebsite}} & \textbf{AP}  & \textbf{Precision} & \textbf{Recall} \\ 
\hline
$\tau = 1$  & 0.69        & 0.73               & 0.63             \\
$\tau = 3$  & {\textbf{0.73}} & {\textbf{0.80}} & \textbf{0.67}          \\

\hline

\hline\\[-3mm]
\end{tabular}
\caption{\textbf{Ablation study of Temporal Window}($\tau$) on NPS~\cite{li2016multi:conf_typical} test set \& AOT~\cite{AOT:IEEEwebsite} val set. Image@1280 reso.}
\label{table:num-of-frames}
\vspace{-4mm}
\end{table}

\subsubsection{Spatial resolution of frame} We find that the spatial resolution of the frame provides a spectrum of operating points to choose the right trade-off between performance and throughput. We use three different resolutions: 1280, 800, and 640. As shown in Table~\ref{tab:resolution}, when we decrease the resolution from 1280 to 640, it only costs \textbf{3\%} absolute drop in AP while gaining \textbf{250\%} in throughput (FPS). 
\begin{table}[h]
\centering
\begin{tabular}{lcccc} 
\hline

\hline

\hline\\[-3mm]
\textbf{Methods} & \textbf{AP}  & \textbf{Precision} & \textbf{Recall} & \textbf{fps}   \\ 
\hline
Image@640         & 0.91          & {0.90}              & 0.84      & \textbf{87.71}       \\
Image@800         & 0.92          & 0.88             & 0.88  & 57.14              \\
Image@1280        & {\textbf{0.95}} & \textbf{0.92}  & \textbf{0.91} & 24.6          \\

\hline

\hline

\hline\\[-3mm]
\end{tabular}
\caption{
\textbf{Ablation of Frame resolution} on NPS~\cite{li2016multi:conf_typical}. 
}
\label{tab:resolution}
\vspace{-2mm}
\end{table}

\subsubsection{Attention Window of VideoSwin}\label{sec:videoswin} We test two configurations of the VideoSwin windowed attention: (1) Attention Window = 3, and Window Shift = 2; and (2) Attention Window = 5, and Window Shift = 0. We keep the spatial size of the window at $8\times8$ and their shift stride at $4\times4$. We vary only the Depth and its corresponding shift stride. We fix $\tau$ to 5 and image resolution to 640 $\times$ 640. As shown in Table~\ref{tab:video-swin}, we can see that results are very close to each other in both performance and throughput. In our best experiment setting, we use attention window = 5, shift = 0 setting due to its higher AP. 
\vspace{-2mm}
\begin{table}[h]
\centering
\begin{tabular}{lcccc} 
\hline

\hline

\hline\\[-3mm]
\textbf{Transformer Settings} & \textbf{AP}  & \textbf{Precision} & \textbf{Recall} & \textbf{FPS}  \\ 
\hline
Att. Window= 3, shift= 2         & 0.914          & 0.875               & 0.83   &   88        \\
Att. Window= 5, shift= 0         & 0.92          & 0.86                 & 0.84   &    87.7         \\

\hline

\hline

\hline\\[-3mm]
\end{tabular}
\caption{\label{tab:video-swin}
\textbf{Ablation study of depth of 3D attention} on NPS dataset~\cite{li2016multi:conf_typical}. Comparison on val data. } 
\vspace{-2mm}
\end{table}

\subsubsection{Effect of Temporal Consistent Augmentation} We train two models on NPS~\cite{li2016multi:conf_typical} at the resolution of 640 $\times$ 640 \&  $\tau=5$. The results (Table \ref{table:TCAstudy}) demonstrate that the proposed Temporal Consistency Augmentation module has better results. It indicates that applied augmentations should be temporally consistent in order to learn robust spatio-temporal features of the drone motion. 

\begin{table}[h]
\centering
\begin{tabular}{lccc} 
\hline

\hline

\hline\\[-3mm]
\textbf{Temporal Augmentations} & \textbf{AP}  & \textbf{Precision} & \textbf{Recall}  \\ 
\hline
inconsistent         & 0.90          & 0.84               & 0.81             \\
consistent         & \textbf{0.92}          & \textbf{0.86}                 & \textbf{0.84}                \\
\hline

\hline

\hline\\[-3mm]
\end{tabular}
\caption{
\label{table:TCAstudy}
\textbf{Ablation study of temporal consistency in augmentations} on NPS ~\cite{li2016multi:conf_typical}. Comparison on val data.} 
\vspace{-2mm}
\end{table}

\section{Tackling real-world drone challenges}
\subsection{Deployment for edge-computing: NVIDIA Jetson Xavier} To show the deployment capability of our model on edge-computing devices, we use NVIDIA Jetson Xavier NX board~\cite{jetson}. 
It has 7025Mb of GPU memory \& 6 CPU cores. Our 640 resolution model obtains the \textbf{real-time fps of 33} without any complex TensorRT optimizations \& keeping the board temperature well below 50$^{\circ}$c.

\subsection{False Positive Per Image (FPPI)}
In real autonomous flight applications, low false positives are required~\cite{AOT:IEEEwebsite} as there is a significant cost attached to flight path maneuvers. FPPI can be calculated by dividing the number of false positives encountered by the number of frames processed. On 171 testing flights (194,193 frames) of the AOT dataset, we obtained the FPPI as low as \textbf{0.000437} vs 0.018~\cite{Ashraf_2021_CVPR:conf_typical}, 0.02474~\cite{dedetr}, 0.0105~\cite{vistr}.

\subsection{Encounter Detection Rate}
To accurately predict the collision path, airborne objects need to be continuously detected and tracked for 3 seconds time within the distance range of 300m~\cite{AOT:IEEEwebsite}. Even though we do not track objects in our method, our model can successfully predict objects within a distance of 700m range. We have continuous detections for 3 seconds in \textbf{82/175} = 46\% closed encounter flights.   

\section{Conclusion}
We have developed an end-to-end trainable and computationally efficient framework for drone-to-drone detection from videos. Spatio-temporal transformer improves drone detection by learning the motion dependencies. Apart from its state-of-the-art performance on three real-world datasets, we have shown that our method significantly improves in throughput as well. We have demonstrated that it is suitable to be deployed on edge-computing devices like NVIDIA Jetson Xavier NX and useful to detect drone encounters (collisions). We will make our code repository publicly available. Since our model is capable of learning spatio-temporal features, in the future, it can be extended to other problems like drone-to-drone tracking or estimating the distance of the airborne objects from a monocular camera.

\bibliographystyle{./IEEEbibstyle/IEEEtran.bst} %
\bibliography{./IEEEbibstyle/IEEEabrv,./IEEEexample}

\begin{thebibliography}{10}
\providecommand{\url}[1]{#1}
\csname url@rmstyle\endcsname
\providecommand{\newblock}{\relax}
\providecommand{\bibinfo}[2]{#2}
\providecommand\BIBentrySTDinterwordspacing{\spaceskip=0pt\relax}
\providecommand\BIBentryALTinterwordstretchfactor{4}
\providecommand\BIBentryALTinterwordspacing{\spaceskip=\fontdimen2\font plus
\BIBentryALTinterwordstretchfactor\fontdimen3\font minus
  \fontdimen4\font\relax}
\providecommand\BIBforeignlanguage[2]{{%
\expandafter\ifx\csname l@#1\endcsname\relax
\typeout{** WARNING: IEEEtran.bst: No hyphenation pattern has been}%
\typeout{** loaded for the language `#1'. Using the pattern for}%
\typeout{** the default language instead.}%
\else
\language=\csname l@#1\endcsname
\fi
#2}}

\bibitem{king2017technology}
A.~King, ``Technology: The future of agriculture,'' \emph{Nature}, vol. 544,
  no. 7651, pp. S21--S23, 2017.

\bibitem{sultani2021human}
W.~Sultani and M.~Shah, ``Human action recognition in drone videos using a few
  aerial training examples,'' \emph{Computer Vision and Image Understanding},
  vol. 206, p. 103186, 2021.

\bibitem{timebalance}
I.~R. Dave, M.~N. Rizve, C.~Chen, and M.~Shah, ``Timebalance:
  Temporally-invariant and temporally-distinctive video representations for
  semi-supervised action recognition,'' in \emph{Proceedings of the IEEE/CVF
  Conference on Computer Vision and Pattern Recognition}, 2023.

\bibitem{gabv1}
M.~N. Rizve, U.~Demir, P.~Tirupattur, A.~J. Rana, K.~Duarte, I.~R. Dave, Y.~S.
  Rawat, and M.~Shah, ``Gabriella: An online system for real-time activity
  detection in untrimmed security videos,'' in \emph{2020 25th International
  Conference on Pattern Recognition (ICPR)}.\hskip 1em plus 0.5em minus
  0.4em\relax IEEE, 2021, pp. 4237--4244.

\bibitem{gabv2}
I.~Dave, Z.~Scheffer, A.~Kumar, S.~Shiraz, Y.~S. Rawat, and M.~Shah,
  ``Gabriellav2: Towards better generalization in surveillance videos for
  action detection,'' in \emph{Proceedings of the IEEE/CVF Winter Conference on
  Applications of Computer Vision}, 2022, pp. 122--132.

\bibitem{spact}
I.~R. Dave, C.~Chen, and M.~Shah, ``Spact: Self-supervised privacy preservation
  for action recognition,'' in \emph{Proceedings of the IEEE/CVF Conference on
  Computer Vision and Pattern Recognition}, 2022, pp. 20\,164--20\,173.

\bibitem{li2021uav}
T.~Li, J.~Liu, W.~Zhang, Y.~Ni, W.~Wang, and Z.~Li, ``Uav-human: A large
  benchmark for human behavior understanding with unmanned aerial vehicles,''
  in \emph{Proceedings of the IEEE/CVF conference on computer vision and
  pattern recognition}, 2021, pp. 16\,266--16\,275.

\bibitem{aakash_iros22}
A.~Kumar, J.~Kini, A.~Mian, and M.~Shah, ``Self supervised learning for
  multiple object tracking in 3d point clouds,'' in \emph{2022 IEEE/RSJ
  International Conference on Intelligent Robots and Systems (IROS)}.\hskip 1em
  plus 0.5em minus 0.4em\relax IEEE, 2022, pp. 3754--3761.

\bibitem{cao2021visdrone}
Y.~Cao, Z.~He, L.~Wang, W.~Wang, Y.~Yuan, D.~Zhang, J.~Zhang, P.~Zhu,
  L.~Van~Gool, J.~Han, \emph{et~al.}, ``Visdrone-det2021: The vision meets
  drone object detection challenge results,'' in \emph{Proceedings of the
  IEEE/CVF International Conference on Computer Vision}, 2021, pp. 2847--2854.

\bibitem{rozantsev2016detecting}
A.~Rozantsev, V.~Lepetit, and P.~Fua, ``Detecting flying objects using a single
  moving camera,'' \emph{IEEE transactions on pattern analysis and machine
  intelligence}, vol.~39, no.~5, pp. 879--892, 2016.

\bibitem{jacobsen2021security}
R.~H. Jacobsen and A.~Marandi, ``Security threats analysis of the unmanned
  aerial vehicle system,'' in \emph{MILCOM 2021-2021 IEEE Military
  Communications Conference (MILCOM)}.\hskip 1em plus 0.5em minus 0.4em\relax
  IEEE, 2021, pp. 316--322.

\bibitem{hassija2021fast}
V.~Hassija, V.~Chamola, A.~Agrawal, A.~Goyal, N.~C. Luong, D.~Niyato, F.~R. Yu,
  and M.~Guizani, ``Fast, reliable, and secure drone communication: A
  comprehensive survey,'' \emph{IEEE Communications Surveys \& Tutorials},
  vol.~23, no.~4, pp. 2802--2832, 2021.

\bibitem{li2016multi:conf_typical}
J.~Li, D.~H. Ye, T.~Chung, M.~Kolsch, J.~Wachs, and C.~Bouman, ``Multi-target
  detection and tracking from a single camera in unmanned aerial vehicles
  (uavs),'' in \emph{2016 IEEE/RSJ international conference on intelligent
  robots and systems (IROS)}.\hskip 1em plus 0.5em minus 0.4em\relax IEEE,
  2016, pp. 4992--4997.

\bibitem{deng2009imagenet}
J.~Deng, W.~Dong, R.~Socher, L.-J. Li, K.~Li, and L.~Fei-Fei, ``Imagenet: A
  large-scale hierarchical image database,'' in \emph{2009 IEEE conference on
  computer vision and pattern recognition}.\hskip 1em plus 0.5em minus
  0.4em\relax Ieee, 2009, pp. 248--255.

\bibitem{coco}
T.-Y. Lin, M.~Maire, S.~Belongie, J.~Hays, P.~Perona, D.~Ramanan,
  P.~Doll{\'a}r, and C.~L. Zitnick, ``Microsoft coco: Common objects in
  context,'' in \emph{European conference on computer vision}.\hskip 1em plus
  0.5em minus 0.4em\relax Springer, 2014, pp. 740--755.

\bibitem{maskrcnn}
K.~He, G.~Gkioxari, P.~Doll{\'a}r, and R.~Girshick, ``Mask r-cnn,'' in
  \emph{Proceedings of the IEEE international conference on computer vision},
  2017, pp. 2961--2969.

\bibitem{yolo}
J.~Redmon, S.~Divvala, R.~Girshick, and A.~Farhadi, ``You only look once:
  Unified, real-time object detection,'' in \emph{Proceedings of the IEEE
  conference on computer vision and pattern recognition}, 2016, pp. 779--788.

\bibitem{dedetr}
X.~Zhu, W.~Su, L.~Lu, B.~Li, X.~Wang, and J.~Dai, ``Deformable detr: Deformable
  transformers for end-to-end object detection,'' \emph{arXiv preprint
  arXiv:2010.04159}, 2020.

\bibitem{Ashraf_2021_CVPR:conf_typical}
M.~W. Ashraf, W.~Sultani, and M.~Shah, ``Dogfight: Detecting drones from drones
  videos,'' in \emph{Proceedings of the IEEE/CVF Conference on Computer Vision
  and Pattern Recognition (CVPR)}, June 2021, pp. 7067--7076.

\bibitem{wang2020cspnet}
C.-Y. Wang, H.-Y.~M. Liao, Y.-H. Wu, P.-Y. Chen, J.-W. Hsieh, and I.-H. Yeh,
  ``Cspnet: A new backbone that can enhance learning capability of cnn,'' in
  \emph{Proceedings of the IEEE/CVF conference on computer vision and pattern
  recognition workshops}, 2020, pp. 390--391.

\bibitem{videoswin}
Z.~Liu, J.~Ning, Y.~Cao, Y.~Wei, Z.~Zhang, S.~Lin, and H.~Hu, ``Video swin
  transformer,'' in \emph{Proceedings of the IEEE/CVF Conference on Computer
  Vision and Pattern Recognition}, 2022, pp. 3202--3211.

\bibitem{8794243}
L.~Dressel and M.~J. Kochenderfer, ``Hunting drones with other drones: Tracking
  a moving radio target,'' in \emph{2019 International Conference on Robotics
  and Automation (ICRA)}, 2019, pp. 1905--1912.

\bibitem{9691910}
S.~Dogru and L.~Marques, ``Drone detection using sparse lidar measurements,''
  \emph{IEEE Robotics and Automation Letters}, vol.~7, no.~2, pp. 3062--3069,
  2022.

\bibitem{rozantsev2016detecting:article_typical}
A.~Rozantsev, V.~Lepetit, and P.~Fua, ``Detecting flying objects using a single
  moving camera,'' \emph{IEEE transactions on pattern analysis and machine
  intelligence}, vol.~39, no.~5, pp. 879--892, 2016.

\bibitem{AOT:IEEEwebsite}
\BIBentryALTinterwordspacing
(2021) The airborne object tracking challenge. [Online]. Available:
  \url{https://www.aicrowd.com/challenges/airborne-object-tracking-challenge}
\BIBentrySTDinterwordspacing

\bibitem{nava2022learning}
M.~Nava, A.~Paolillo, J.~Guzzi, L.~M. Gambardella, and A.~Giusti, ``Learning
  visual localization of a quadrotor using its noise as self-supervision,''
  \emph{IEEE Robotics and Automation Letters}, vol.~7, no.~2, pp. 2218--2225,
  2022.

\bibitem{li2022self}
S.~Li, C.~De~Wagter, and G.~C. De~Croon, ``Self-supervised monocular
  multi-robot relative localization with efficient deep neural networks,'' in
  \emph{2022 International Conference on Robotics and Automation (ICRA)}.\hskip
  1em plus 0.5em minus 0.4em\relax IEEE, 2022, pp. 9689--9695.

\bibitem{inproceedings}
K.~Yang and Q.~Quan, ``An autonomous intercept drone with image-based visual
  servo,'' 05 2020, pp. 2230--2236.

\bibitem{9561365}
F.~Chen, Y.~Lu, Y.~Li, and X.~Xie, ``Real-time active detection of targets and
  path planning using uavs,'' in \emph{2021 IEEE International Conference on
  Robotics and Automation (ICRA)}, 2021, pp. 391--397.

\bibitem{unknown}
Z.~Cao, C.~Fu, J.~Ye, B.~Li, and Y.~Li, ``Siamapn++: Siamese attentional
  aggregation network for real-time uav tracking,'' 06 2021.

\bibitem{9341432}
H.~Yu, F.~Zhang, P.~Huang, C.~Wang, and L.~Yuanhao, ``Autonomous obstacle
  avoidance for uav based on fusion of radar and monocular camera,'' in
  \emph{2020 IEEE/RSJ International Conference on Intelligent Robots and
  Systems (IROS)}, 2020, pp. 5954--5961.

\bibitem{vtn}
D.~Neimark, O.~Bar, M.~Zohar, and D.~Asselmann, ``Video transformer network,''
  in \emph{Proceedings of the IEEE/CVF International Conference on Computer
  Vision (ICCV) Workshops}, October 2021, pp. 3163--3172.

\bibitem{swin}
Z.~Liu, Y.~Lin, Y.~Cao, H.~Hu, Y.~Wei, Z.~Zhang, S.~Lin, and B.~Guo, ``Swin
  transformer: Hierarchical vision transformer using shifted windows,'' in
  \emph{Proceedings of the IEEE/CVF International Conference on Computer
  Vision}, 2021, pp. 10\,012--10\,022.

\bibitem{tclr}
\BIBentryALTinterwordspacing
I.~Dave, R.~Gupta, M.~N. Rizve, and M.~Shah, ``Tclr: Temporal contrastive
  learning for video representation,'' \emph{Computer Vision and Image
  Understanding}, p. 103406, 2022. [Online]. Available:
  \url{https://www.sciencedirect.com/science/article/pii/S1077314222000376}
\BIBentrySTDinterwordspacing

\bibitem{svt}
K.~Ranasinghe, M.~Naseer, S.~Khan, F.~S. Khan, and M.~S. Ryoo,
  ``Self-supervised video transformer,'' in \emph{Proceedings of the IEEE/CVF
  Conference on Computer Vision and Pattern Recognition}, 2022, pp. 2874--2884.

\bibitem{yolov4}
A.~Bochkovskiy, C.-Y. Wang, and H.-Y.~M. Liao, ``Yolov4: Optimal speed and
  accuracy of object detection,'' \emph{arXiv preprint arXiv:2004.10934}, 2020.

\bibitem{spp}
K.~He, X.~Zhang, S.~Ren, and J.~Sun, ``Spatial pyramid pooling in deep
  convolutional networks for visual recognition,'' \emph{IEEE transactions on
  pattern analysis and machine intelligence}, vol.~37, no.~9, pp. 1904--1916,
  2015.

\bibitem{NIPS2017_3f5ee243}
\BIBentryALTinterwordspacing
A.~Vaswani, N.~Shazeer, N.~Parmar, J.~Uszkoreit, L.~Jones, A.~N. Gomez, L.~u.
  Kaiser, and I.~Polosukhin, ``Attention is all you need,'' in \emph{Advances
  in Neural Information Processing Systems}, I.~Guyon, U.~V. Luxburg,
  S.~Bengio, H.~Wallach, R.~Fergus, S.~Vishwanathan, and R.~Garnett, Eds.,
  vol.~30.\hskip 1em plus 0.5em minus 0.4em\relax Curran Associates, Inc.,
  2017. [Online]. Available:
  \url{https://proceedings.neurips.cc/paper/2017/file/3f5ee243547dee91fbd053c1c4a845aa-Paper.pdf}
\BIBentrySTDinterwordspacing

\bibitem{Zhu_2021_ICCV}
X.~Zhu, S.~Lyu, X.~Wang, and Q.~Zhao, ``Tph-yolov5: Improved yolov5 based on
  transformer prediction head for object detection on drone-captured
  scenarios,'' in \emph{Proceedings of the IEEE/CVF International Conference on
  Computer Vision (ICCV) Workshops}, October 2021, pp. 2778--2788.

\bibitem{kingma2014adam:article_typical}
D.~P. Kingma and J.~Ba, ``Adam: A method for stochastic optimization,''
  \emph{arXiv preprint arXiv:1412.6980}, 2014.

\bibitem{https://doi.org/10.48550/arxiv.1812.01187}
\BIBentryALTinterwordspacing
T.~He, Z.~Zhang, H.~Zhang, Z.~Zhang, J.~Xie, and M.~Li, ``Bag of tricks for
  image classification with convolutional neural networks,'' 2018. [Online].
  Available: \url{https://arxiv.org/abs/1812.01187}
\BIBentrySTDinterwordspacing

\bibitem{glennjocher20204154370}
\BIBentryALTinterwordspacing
G.~Jocher, A.~Stoken, J.~Borovec, NanoCode012, ChristopherSTAN, L.~Changyu,
  Laughing, tkianai, A.~Hogan, lorenzomammana, yxNONG, AlexWang1900,
  L.~Diaconu, Marc, wanghaoyang0106, ml5ah, Doug, F.~Ingham, Frederik, Guilhen,
  Hatovix, J.~Poznanski, J.~Fang, L.~Yu, changyu98, M.~Wang, N.~Gupta,
  O.~Akhtar, PetrDvoracek, and P.~Rai. (2020, Oct.) {ultralytics/yolov5: v3.1 -
  Bug Fixes and Performance Improvements}. [Online]. Available:
  \url{https://doi.org/10.5281/zenodo.4154370}
\BIBentrySTDinterwordspacing

\bibitem{lin2014microsoft:conf_typical}
T.-Y. Lin, M.~Maire, S.~Belongie, J.~Hays, P.~Perona, D.~Ramanan,
  P.~Doll{\'a}r, and C.~L. Zitnick, ``Microsoft coco: Common objects in
  context,'' in \emph{European conference on computer vision}.\hskip 1em plus
  0.5em minus 0.4em\relax Springer, 2014, pp. 740--755.

\bibitem{ICCVWorkshop:IEEEwebsite}
\BIBentryALTinterwordspacing
(2021) The airborne object tracking challenge. [Online]. Available:
  \url{https://zontakm9.github.io/aot-iccvw21/}
\BIBentrySTDinterwordspacing

\bibitem{scrdet}
X.~Yang, J.~Yang, J.~Yan, Y.~Zhang, T.~Zhang, Z.~Guo, X.~Sun, and K.~Fu,
  ``Scrdet: Towards more robust detection for small, cluttered and rotated
  objects,'' in \emph{Proceedings of the IEEE/CVF International Conference on
  Computer Vision}, 2019, pp. 8232--8241.

\bibitem{fcos}
Z.~Tian, C.~Shen, H.~Chen, and T.~He, ``Fcos: Fully convolutional one-stage
  object detection,'' in \emph{Proceedings of the IEEE/CVF international
  conference on computer vision}, 2019, pp. 9627--9636.

\bibitem{mega}
Y.~Chen, Y.~Cao, H.~Hu, and L.~Wang, ``Memory enhanced global-local aggregation
  for video object detection,'' in \emph{Proceedings of the IEEE/CVF conference
  on computer vision and pattern recognition}, 2020, pp. 10\,337--10\,346.

\bibitem{slsa}
H.~Wu, Y.~Chen, N.~Wang, and Z.~Zhang, ``Sequence level semantics aggregation
  for video object detection,'' in \emph{Proceedings of the IEEE/CVF
  international conference on computer vision}, 2019, pp. 9217--9225.

\bibitem{vistr}
Y.~Wang, Z.~Xu, X.~Wang, C.~Shen, B.~Cheng, H.~Shen, and H.~Xia, ``End-to-end
  video instance segmentation with transformers,'' in \emph{Proceedings of the
  IEEE/CVF Conference on Computer Vision and Pattern Recognition}, 2021, pp.
  8741--8750.

\bibitem{jetson}
\BIBentryALTinterwordspacing
(2021) Nvidia jetson xavier nx. [Online]. Available:
  \url{https://www.nvidia.com/en-us/autonomous-machines/embedded-systems/jetson-xavier-nx/}
\BIBentrySTDinterwordspacing

\end{thebibliography}

\end{document}